\documentclass{article}

\usepackage{times}
\usepackage{graphicx} 
\graphicspath{ {./} }
\usepackage{subfigure} 

\usepackage{natbib}

\usepackage{algorithm}
\usepackage{algorithmic}

\usepackage{hyperref}

\usepackage[accepted]{icml2017}

\icmltitlerunning{Generating Reviews and Discovering Sentiment}

\begin{document} 

\twocolumn[
\icmltitle{Learning to Generate Reviews and Discovering Sentiment}

\icmlsetsymbol{equal}{*}

\begin{icmlauthorlist}
\icmlauthor{Alec Radford}{to}
\icmlauthor{Rafal Jozefowicz}{to}
\icmlauthor{Ilya Sutskever}{to}
\end{icmlauthorlist}

\icmlaffiliation{to}{OpenAI, San Francisco, California, USA}

\icmlcorrespondingauthor{Alec Radford}{alec@openai.com}

\icmlkeywords{machine learning}

\vskip 0.3in
]

\printAffiliationsAndNotice{}

\begin{abstract}
We explore the properties of byte-level recurrent language models. When given sufficient amounts of capacity, training data, and compute time, the representations learned by these models include disentangled features corresponding to high-level concepts. Specifically, we find a single unit which performs sentiment analysis. These representations, learned in an unsupervised manner, achieve state of the art on the binary subset of the Stanford Sentiment Treebank. They are also very data efficient. When using only a handful of labeled examples, our approach matches the performance of strong baselines trained on full datasets. We also demonstrate the sentiment unit has a direct influence on the generative process of the model. Simply fixing its value to be positive or negative generates samples with the corresponding positive or negative sentiment.
\end{abstract}

\section{Introduction and Motivating Work}
Representation learning \cite{bengio2013representation} plays a critical role in many modern machine learning systems. Representations map raw data to more useful forms and the choice of representation is an important component of any application. Broadly speaking, there are two areas of research emphasizing different details of how to learn useful representations. 

The supervised training of high-capacity models on large labeled datasets is critical to the recent success of deep learning techniques for a wide range of applications such as image classification \cite{krizhevsky2012imagenet}, speech recognition \cite{hinton2012deep}, and machine translation \cite{wu2016google}. Analysis of the task specific representations learned by these models reveals many fascinating properties \cite{zhou2014object}. Image classifiers learn a broadly useful hierarchy of feature detectors re-representing raw pixels as edges, textures, and objects \cite{zeiler2014visualizing}. In the field of computer vision, it is now commonplace to reuse these representations on a broad suite of related tasks - one of the most successful examples of transfer learning to date \cite{oquab2014learning}.

There is also a long history of unsupervised representation learning \cite{olshausen1997sparse}. Much of the early research into modern deep learning was developed and validated via this approach \cite{hinton2006reducing} \cite{huang2007unsupervised} \cite{vincent2008extracting} \cite{coates2010analysis} \cite{le2013building}. Unsupervised learning is promising due to its ability to scale beyond only the subsets and domains of data that can be cleaned and labeled given resource, privacy, or other constraints. This advantage is also its difficulty. While supervised approaches have clear objectives that can be directly optimized, unsupervised approaches rely on proxy tasks such as reconstruction, density estimation, or generation, which do not directly encourage useful representations for specific tasks. As a result, much work has gone into designing objectives, priors, and architectures meant to encourage the learning of useful representations. We refer readers to \citet{goodfellow2016deep} for a detailed review.

Despite these difficulties, there are notable applications of unsupervised learning. Pre-trained word vectors are a vital part of many modern NLP systems \cite{collobert2011natural}. These representations, learned by modeling word co-occurrences, increase the data efficiency and generalization capability of NLP systems \cite{pennington2014glove} \cite{chen2014fast}. Topic modelling can also discover factors within a corpus of text which align to human interpretable concepts such as “art” or “education” \cite{blei2003latent}.

How to learn representations of phrases, sentences, and documents is an open area of research. Inspired by the success of word vectors, \citet{kiros2015skip} propose skip-thought vectors, a method of training a sentence encoder by predicting the preceding and following sentence. The representation learned by this objective performs competitively on a broad suite of evaluated tasks. More advanced training techniques such as layer normalization \cite{ba2016layer} further improve results. However, skip-thought vectors are still outperformed by supervised models which directly optimize the desired performance metric on a specific dataset. This is the case for both text classification tasks, which measure whether a specific concept is well encoded in a representation, and more general semantic similarity tasks. This occurs even when the datasets are relatively small by modern standards, often consisting of only a few thousand labeled examples. 

In contrast to learning a generic representation on one large dataset and then evaluating on other tasks/datasets, \citet{dai2015semi} proposed using similar unsupervised objectives such as sequence autoencoding and language modeling to first pretrain a model on a dataset and then finetune it for a given task. This approach outperformed training the same model from random initialization and achieved state of the art on several text classification datasets. Combining language modelling with topic modelling and fitting a small supervised feature extractor on top has also achieved strong results on in-domain document level sentiment analysis \cite{dieng2016topicrnn}.

Considering this, we hypothesize two effects may be combining to result in the weaker performance of purely unsupervised approaches. Skip-thought vectors were trained on a corpus of books. But some of the classification tasks they are evaluated on, such as sentiment analysis of reviews of consumer goods, do not have much overlap with the text of novels. We propose this distributional issue, combined with the limited capacity of current models, results in representational underfitting. Current generic distributed sentence representations may be very lossy - good at capturing the gist, but poor with the precise semantic or syntactic details which are critical for applications.

The experimental and evaluation protocols may be underestimating the quality of unsupervised representation learning for sentences and documents due to certain seemingly insignificant design decisions. \citet{hill2016learning} also raises concern about current evaluation tasks in their recent work which provides a thorough survey of architectures and objectives for learning unsupervised sentence representations - including the above mentioned skip-thoughts.

In this work, we test whether this is the case. We focus in on the task of sentiment analysis and attempt to learn an unsupervised representation that accurately contains this concept. \citet{mikolov2013linguistic} showed that word-level recurrent language modelling supports the learning of useful word vectors and we are interested in pushing this line of work. As an approach, we consider the popular research benchmark of byte (character) level language modelling due to its further simplicity and generality. We are also interested in evaluating this approach as it is not immediately clear whether such a low-level training objective supports the learning of high-level representations. We train on a very large corpus picked to have a similar distribution as our task of interest. We also benchmark on a wider range of tasks to quantify the sensitivity of the learned representation to various degrees of out-of-domain data and tasks.

\section{Dataset}
Much previous work on language modeling has evaluated on relatively small but competitive datasets such as Penn Treebank \cite{marcus1993building} and Hutter Prize Wikipedia \cite{hutterhuman}. As discussed in \citet{jozefowicz2016exploring} performance on these datasets is primarily dominated by regularization. Since we are interested in high-quality sentiment representations, we chose the Amazon product review dataset introduced in \citet{mcauley2015inferring} as a training corpus. In de-duplicated form, this dataset contains over 82 million product reviews from May 1996 to July 2014 amounting to over 38 billion training bytes. Due to the size of the dataset, we first split it into 1000 shards containing equal numbers of reviews and set aside 1 shard for validation and 1 shard for test.

\begin{figure}[h]
\vskip 0.2in
\begin{center}
\centerline{\includegraphics[width=\columnwidth]{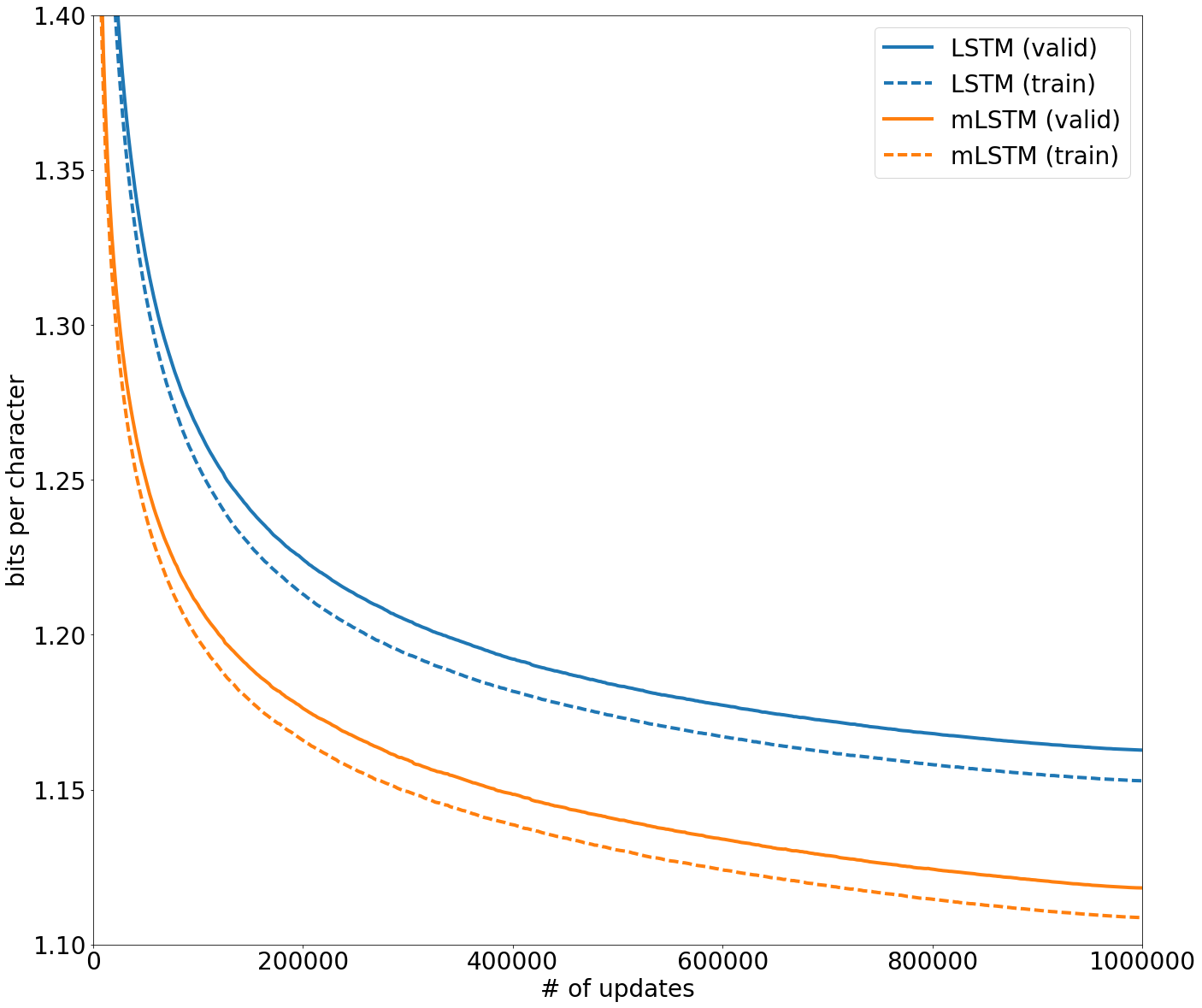}}
\caption{The mLSTM converges faster and achieves a better result within our time budget compared to a standard LSTM with the same hidden state size}
\label{convergence comparison}
\end{center}
\vskip -0.2in
\end{figure} 

\section{Model and Training Details}
Many potential recurrent architectures and hyperparameter settings were considered in preliminary experiments on the dataset. Given the size of the dataset, searching the wide space of possible configurations is quite costly. To help alleviate this, we evaluated the generative performance of smaller candidate models after a single pass through the dataset. The model chosen for the large scale experiment is a single layer multiplicative LSTM \cite{krause2016multiplicative} with 4096 units. We observed multiplicative LSTMs to converge faster than normal LSTMs for the hyperparameter settings that were explored both in terms of data and wall-clock time. The model was trained for a single epoch on mini-batches of 128 subsequences of length 256 for a total of ~1 million weight updates. States were initialized to zero at the beginning of each shard and persisted across updates to simulate full-backpropagation and allow for the forward propagation of information outside of a given subsequence. Adam \cite{kingma2014adam} was used to accelerate learning with an initial 5e-4 learning rate that was decayed linearly to zero over the course of training. Weight normalization \cite{salimans2016weight} was applied to the LSTM parameters. Data-parallelism was used across 4 Pascal Titan X gpus to speed up training and increase effective memory size. Training took approximately one month. The model is compact, containing approximately as many parameters as there are reviews in the training dataset. It also has a high ratio of compute to total parameters compared to other large scale language models due to operating at a byte level. The selected model reaches 1.12 bits per byte.

\begin{figure}[h]
\vskip 0.2in
\begin{center}
\centerline{\includegraphics[width=\columnwidth]{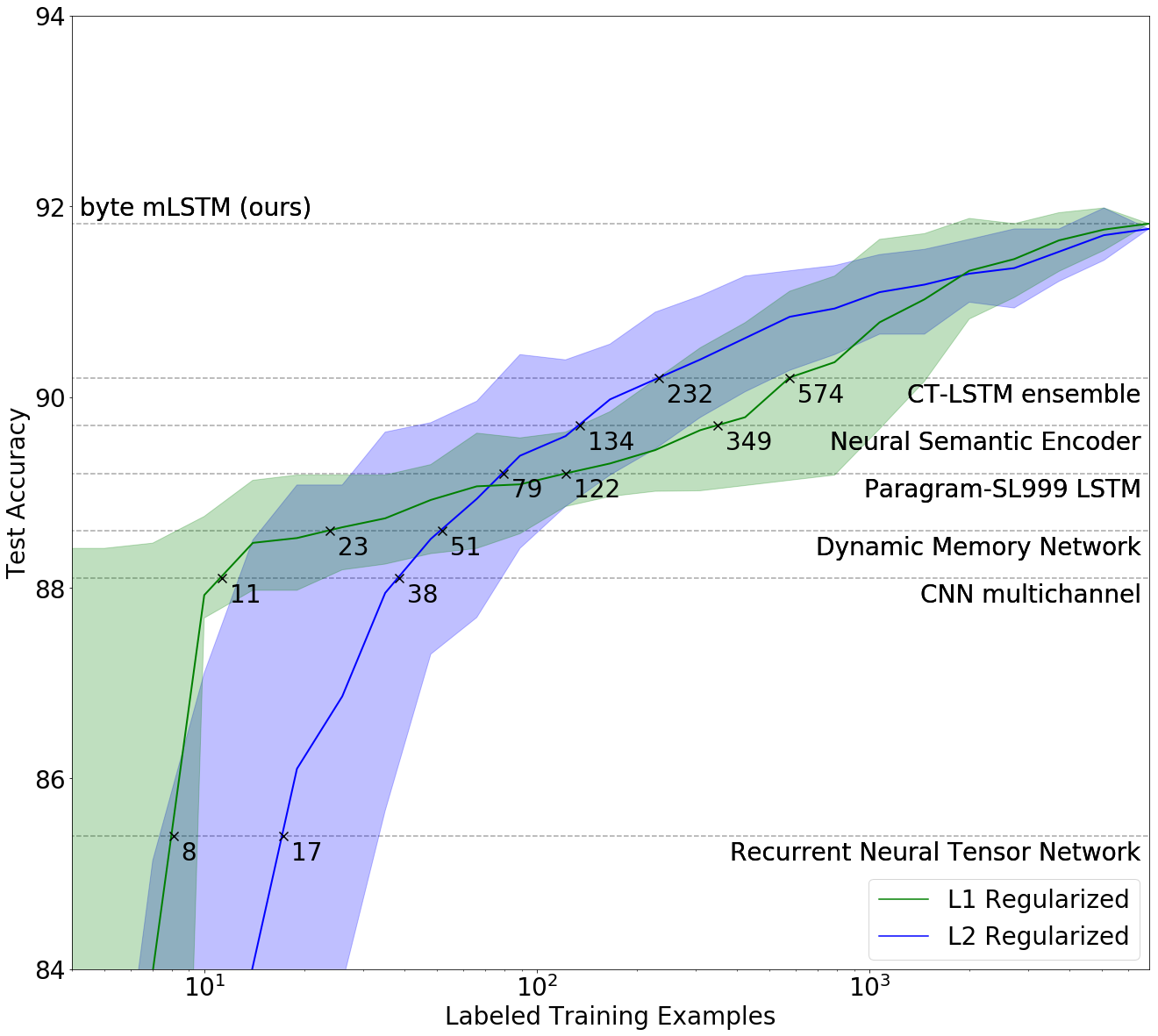}}
\caption{Performance on the binary version of SST as a function of labeled training examples. The solid lines indicate the average of 100 runs while the sharded regions indicate the 10th and 90th percentiles. Previous results on the dataset are plotted as dashed lines with the numbers indicating the amount of examples required for logistic regression on the byte mLSTM representation to match their performance. RNTN \cite{socher2013recursive} CNN \cite{kim2014convolutional} DMN \cite{kumar2015ask} LSTM \cite{wieting2015towards} NSE \cite{munkhdalai2016neural} CT-LSTM \cite{looks2017deep}}
\label{sst binary performance}
\end{center}
\vskip -0.2in
\end{figure} 

\begin{table}[h]
\caption{Small dataset classification accuracies}
\label{small-classification}
\vskip 0.15in
\begin{center}
\begin{small}
\begin{sc}
\begin{tabular}{lccccc}
\hline
\abovespace\belowspace
Method & MR & CR & SUBJ & MPQA\\
\hline
\abovespace
NBSVM [\citenum{wang2012baselines}] & 79.4& 81.8& 93.2& 86.3\\
\hline
SkipThought [\citenum{kiros2015skip}] & 77.3& 81.8& 92.6& 87.9\\
SkipThought(LN)& 79.5& 83.1& 93.7& 89.3\\
SDAE [\citenum{hill2016learning}] & 74.6& 78.0& 90.8& 86.9\\
\hline
Cnn [\citenum{kim2014convolutional}] & 81.5& 85.0& 93.4& 89.6\\
Adasent [\citenum{zhao2015self}] & 83.1& 86.3& \textbf{95.5}& \textbf{93.3}\\
\hline
byte mLSTM     & \textbf{86.9}& \textbf{91.4}& 94.6& 88.5\\
\hline
\end{tabular}
\end{sc}
\end{small}
\end{center}
\vskip -0.1in
\end{table}

\section{Experimental Setup and Results}

Our model processes text as a sequence of UTF-8 encoded bytes \cite{yergeau2003utf}. For each byte, the model updates its hidden state and predicts a probability distribution over the next possible byte. The hidden state of the model serves as an online summary of the sequence which encodes all information the model has learned to preserve that is relevant to predicting the future bytes of the sequence. We are interested in understanding the properties of the learned encoding. The process of extracting a feature representation is outlined as follows:

\begin{itemize}
\item Since newlines are used as review delimiters in the training dataset, all newline characters are replaced with spaces to avoid the model resetting state.
\item Any leading whitespace is removed and replaced with a newline+space to simulate a start token. Any trailing whitespace is removed and replaced with a space to simulate an end token. The text is encoded as a UTF-8 byte sequence.
\item Model states are initialized to zeros. The model processes the sequence and the final cell states of the mLSTM are used as a feature representation. Tanh is applied to bound values between -1 and 1.
\end{itemize}

We follow the methodology established in \citet{kiros2015skip} by training a logistic regression classifier on top of our model's representation on datasets for tasks including semantic relatedness, text classification, and paraphrase detection. For the details on these comparison experiments, we refer the reader to their work. One exception is that we use an L1 penalty for text classification results instead of L2 as we found this performed better in the very low data regime.

\subsection{Review Sentiment Analysis}

Table 1 shows the results of our model on 4 standard text classification datasets. The performance of our model is noticeably lopsided. On the MR \cite{pang2005seeing} and CR \cite{hu2004mining} sentiment analysis datasets we improve the state of the art by a significant margin. The MR and CR datasets are sentences extracted from Rotten Tomatoes, a movie review website, and Amazon product reviews (which almost certainly overlaps with our training corpus). This suggests that our model has learned a rich representation of text from a similar domain. On the other two datasets, SUBJ's subjectivity/objectivity detection \cite{pang2004sentimental} and MPQA's opinion polarity \cite{wiebe2005annotating} our model has no noticeable advantage over other unsupervised representation learning approaches and is still outperformed by a supervised approach.

To better quantify the learned representation, we also test on a wider set of sentiment analysis datasets with different properties. The Stanford Sentiment Treebank (SST) \cite{socher2013recursive} was created specifically to evaluate more complex compositional models of language. It is derived from the same base dataset as MR but was relabeled via Amazon Mechanical and includes dense labeling of the phrases of parse trees computed for all sentences. For the binary subtask, this amounts to 76961 total labels compared to the 6920 sentence level labels. As a demonstration of the capability of unsupervised representation learning to simplify data collection and remove preprocessing steps, our reported results ignore these dense labels and computed parse trees, using only the raw text and sentence level labels.

The representation learned by our model achieves 91.8\% significantly outperforming the state of the art of 90.2\% by a 30 model ensemble \cite{looks2017deep}. As visualized in Figure 2, our model is very data efficient. It matches the performance of baselines using as few as a dozen labeled examples and outperforms all previous results with only a few hundred labeled examples. This is under 10\% of the total sentences in the dataset. Confusingly, despite a 16\% relative error reduction on the binary subtask, it does not reach the state of the art of 53.6\% on the fine-grained subtask, achieving 52.9\%. 

\begin{figure}[h]
\vskip 0.2in
\begin{center}
\centerline{\includegraphics[width=\columnwidth]{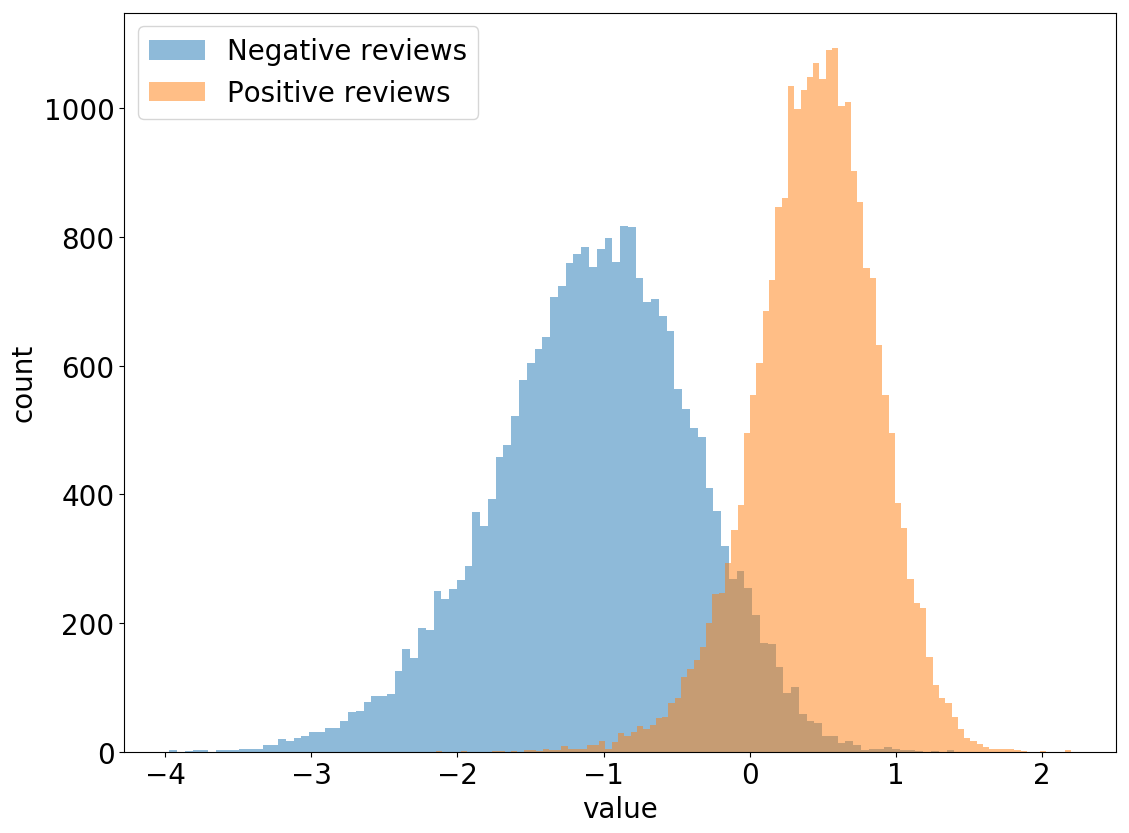}}
\caption{Histogram of cell activation values for the sentiment unit on IMDB reviews.}
\label{activation histogram}
\end{center}
\vskip -0.2in
\end{figure} 

\begin{figure}[h]
\vskip 0.2in
\begin{center}
\centerline{\includegraphics[width=\columnwidth]{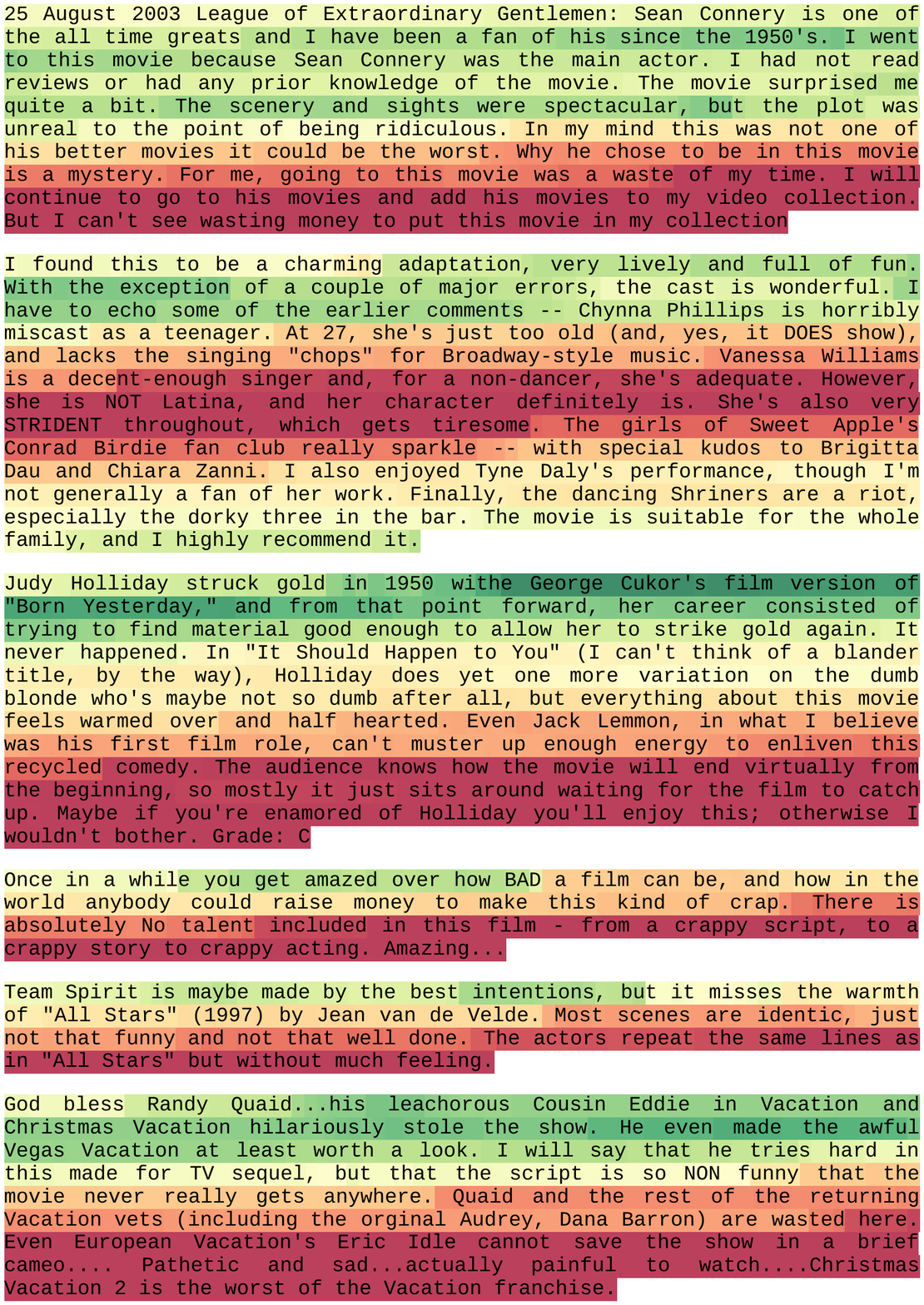}}
\caption{Visualizing the value of the sentiment cell as it processes six randomly selected high contrast IMDB reviews. Red indicates negative sentiment while green indicates positive sentiment. Best seen in color.}
\label{activation histogram}
\end{center}
\vskip -0.2in
\end{figure} 

\subsection{Sentiment Unit}

\begin{table}[h]
\caption{IMDB sentiment classification}
\label{imdb-sentiment}
\vskip 0.15in
\begin{center}
\begin{small}
\begin{sc}
\begin{tabular}{lc}
\hline
\abovespace\belowspace
Method & Error\\
\hline
\abovespace
FullUnlabeledBoW \cite{maas2011learning} & 11.11\%\\
NB-SVM trigram \cite{mesnil2014ensemble} & 8.13\%\\
\textbf{Sentiment unit (ours)} & 7.70\%\\
SA-LSTM \cite{dai2015semi} & 7.24\%\\
\textbf{byte mLSTM (ours)} & 7.12\%\\
TopicRNN \cite{dieng2016topicrnn} & 6.24\%\\
Virtual Adv \cite{miyato2016adversarial} & 5.91\%\\
\hline
\end{tabular}
\end{sc}
\end{small}
\end{center}
\vskip -0.1in
\end{table}

We conducted further analysis to understand what representations our model learned and how they achieve the observed data efficiency. The benefit of an L1 penalty in the low data regime (see Figure 2) is a clue. L1 regularization is known to reduce sample complexity when there are many irrelevant features \cite{ng2004feature}. This is likely to be the case for our model since it is trained as a language model and not as a supervised feature extractor. By inspecting the relative contributions of features on various datasets, we discovered a single unit within the mLSTM \textit{that directly corresponds to sentiment}. In Figure 3 we show the histogram of the final activations of this unit after processing IMDB reviews \cite{maas2011learning} which shows a bimodal distribution with a clear separation between positive and negative reviews. In Figure 4 we visualize the activations of this unit on 6 randomly selected reviews from a set of 100 high contrast reviews which shows it acts as an online estimate of the local sentiment of the review. Fitting a threshold to this \textit{single} unit achieves a test accuracy of 92.30\% which outperforms a strong supervised results on the dataset, the 91.87\% of NB-SVM trigram \cite{mesnil2014ensemble}, but is still below the semi-supervised state of the art of 94.09\% \cite{miyato2016adversarial}. Using the full 4096 unit representation achieves 92.88\%. This is an improvement of only 0.58\% over the sentiment unit suggesting that almost all information the model retains that is relevant to sentiment analysis is represented in the very compact form of a single scalar. Table 2 has a full list of results on the IMDB dataset.

\begin{figure}[h]
\vskip 0.2in
\begin{center}
\centerline{\includegraphics[width=\columnwidth]{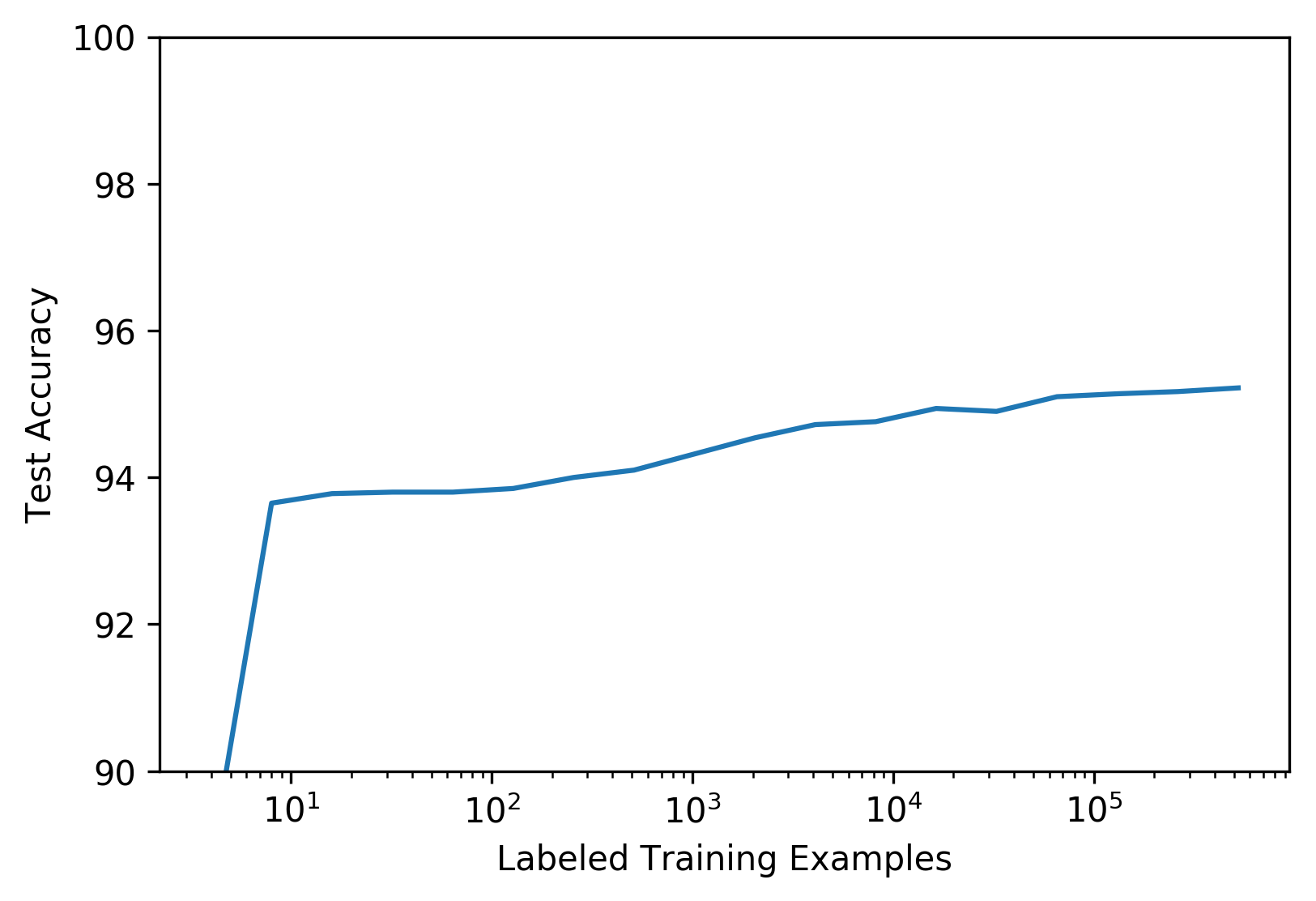}}
\caption{Performance on the binary version of the Yelp reviews dataset as a function of labeled training examples. The model's performance plateaus after about ten labeled examples and only slow improves with additional data.}
\label{capacity ceiling}
\end{center}
\vskip -0.2in
\end{figure} 

\begin{table}[h]
\caption{Microsoft Paraphrase Corpus}
\label{paraphrase}
\vskip 0.15in
\begin{center}
\begin{small}
\begin{sc}
\begin{tabular}{lccc}
\hline
\abovespace\belowspace
Method & Acc & F1\\
\hline
\abovespace
SkipThought \cite{kiros2015skip} & 73.0& 82.0\\
SDAE \cite{hill2016learning} & 76.4& 83.4\\
\hline
MTMETRICS [\citenum{madnani2012re}] & \textbf{77.4}  & \textbf{84.1}\\
\hline
byte mLSTM     & 75.0& 82.8&\\
\hline
\end{tabular}
\end{sc}
\end{small}
\end{center}
\vskip -0.1in
\end{table}

\begin{table}[!h]
\caption{SICK semantic relatedness subtask}
\label{SICK}
\vskip 0.15in
\begin{center}
\begin{small}
\begin{sc}
\begin{tabular}{lcccc}
\hline
\abovespace\belowspace
Method & \textit{r} & $\rho$ & \textbf{MSE}\\
\hline
\abovespace
SkipThought [\citenum{kiros2015skip}] & 0.858& 0.792& 0.269\\
SkipThought(LN)& 0.858& 0.788& 0.270\\
\hline
Tree-LSTM [\citenum{tai2015improved}] & \textbf{0.868}& \textbf{0.808}& \textbf{0.253}\\
\hline
byte mLSTM     & 0.792& 0.725& 0.390\\
\hline
\end{tabular}
\end{sc}
\end{small}
\end{center}
\vskip -0.1in
\end{table}

\begin{table*}[!h]
\begin{center}
\label{sentiment-samples}
\def\arraystretch{1.5}
\begin{tabular}{p{8cm}p{8cm}}
\hline
{\bf Sentiment fixed to positive} & {\bf Sentiment fixed to negative} \\
\hline
Just what I was looking for. Nice fitted pants, exactly matched seam to color contrast with other pants I own. Highly recommended and also very happy! & The package received was blank and has no barcode. A waste of time and money. \\ 
This product does what it is supposed to.  I always keep three of these in my kitchen just in case ever I need a replacement cord. & Great little item. Hard to put on the crib without some kind of embellishment. My guess is just like the screw kind of attachment I had. \\
Best hammock ever! Stays in place and holds it's shape. Comfy (I love the deep neon pictures on it), and looks so cute. & They didn't fit either. Straight high sticks at the end.  On par with other buds I have.  Lesson learned to avoid. \\
Dixie is getting her Doolittle newsletter we'll see another new one coming out next year.  Great stuff.  And, here's the contents -  information that we hardly know about or forget. & great product but no seller. couldn't ascertain a cause. Broken product. I am a prolific consumer of this company all the time. \\
I love this weapons look . Like I said beautiful !!! I recommend it to all. Would suggest this to many roleplayers , And I stronge to get them for every one I know. A must watch for any man who love Chess! & Like the cover, Fits good. . However, an annoying rear piece like garbage should be out of this one. I bought this hoping it would help with a huge pull down my back \&  the black just doesn't stay. Scrap off everytime I use it.... Very disappointed. \\
\end{tabular}
\vspace{-4pt}
\caption{Random samples from the model generated when the value of sentiment hidden state is fixed to either -1 or 1 for all steps. The sentiment unit has a strong influence on the model's generative process.}
\end{center}
\end{table*}

\subsection{Capacity Ceiling}

Encouraged by these results, we were curious how well the model's representation scales to larger datasets. We try our approach on the binary version of the Yelp Dataset Challenge in 2015 as introduced in \citet{zhang2015character}. This dataset contains 598,000 examples which is an order of magnitude larger than any other datasets we tested on. When visualizing performance as a function of number of training examples in Figure 5, we observe a "capacity ceiling" where the test accuracy of our approach only improves by a little over 1\% across a four order of magnitude increase in training data. Using the full dataset, we achieve 95.22\% test accuracy. This better than a BoW TFIDF baseline at 93.66\% but slightly worse than the 95.64\% of a linear classifier on top of the 500,000 most frequent n-grams up to length 5.

The observed capacity ceiling is an interesting phenomena and stumbling point for scaling our unsupervised representations. We think a variety of factors are contributing to cause this. Since our model is trained only on Amazon reviews, it is does not appear to be sensitive to concepts specific to other domains. For instance, Yelp reviews are of businesses, where details like hospitality, location, and atmosphere are important. But these ideas are not present in reviews of products. Additionally, there is a notable drop in the relative performance of our approach transitioning from sentence to document datasets. This is likely due to our model working on the byte level which leads to it focusing on the content of the last few sentences instead of the whole document. Finally, as the amount of labeled data increases, the performance of the simple linear model we train on top of our static representation will eventually saturate. Complex models explicitly trained for a task can continue to improve and eventually outperform our approach with enough labeled data.

With this context, the observed results make a lot of sense. On a small sentence level dataset of a known domain (the movie reviews of Stanford Sentiment Treebank) our model sets a new state of the art. But on a large, document level dataset of a different domain (the Yelp reviews) it is only competitive with standard baselines.

\subsection{Other Tasks}

Besides classification, we also evaluate on two other standard tasks: semantic relatedness and paraphrase detection. While our model performs competitively on Microsoft Research Paraphrase Corpus \cite{dolan2004unsupervised} in Table 3, it performs poorly on the SICK semantic relatedness task \cite{marelli2014semeval} in Table 4. It is likely that the form and content of the semantic relatedness task, which is built on top of descriptions of images and videos and contains sentences such as "A sea turtle is hunting for fish" is effectively out-of-domain for our model which has only been trained on the text of product reviews.

\subsection{Generative Analysis}

Although the focus of our analysis has been on the properties of our model's representation, it is trained as a generative model and we are also interested in its generative capabilities. \citet{hu2017controllable} and \citet{dong2017learning} both designed conditional generative models to disentangle the content of text from various attributes like sentiment or tense. We were curious whether a similar result could be achieved using the sentiment unit. In Table 5 we show that by simply setting the sentiment unit to be positive or negative, the model generates corresponding positive or negative reviews. While all sampled negative reviews contain sentences with negative sentiment, they sometimes contain sentences with positive sentiment as well. This might be reflective of the bias of the training corpus which contains over 5x as many five star reviews as one star reviews. Nevertheless, it is interesting to see that such a simple manipulation of the model's representation has a noticeable effect on its behavior. The samples are also high quality for a byte level language model and often include valid sentences.

\section{Discussion and Future Work}
It is an open question why our model recovers the concept of sentiment in such a precise, disentangled, interpretable, and manipulable way. It is possible that sentiment as a conditioning feature has strong predictive capability for language modelling. This is likely since sentiment is such an important component of a review. Previous work analysing LSTM language models showed the existence of interpretable units that indicate position within a line or presence inside a quotation \cite{karpathy2015visualizing}. In many ways, the sentiment unit in this model is just a scaled up example of the same phenomena. The update equation of an LSTM could play a role. The element-wise operation of its gates may encourage axis-aligned representations. Models such as word2vec have also been observed to have small subsets of dimensions strongly associated with specific tasks \cite{li2016understanding}. 

Our work highlights the sensitivity of learned representations to the data distribution they are trained on. The results make clear that it is unrealistic to expect a model trained on a corpus of books, where the two most common genres are Romance and Fantasy, to learn an encoding which preserves the exact sentiment of a review. Likewise, it is unrealistic to expect a model trained on Amazon product reviews to represent the precise semantic content of a caption of an image or a video.

There are several promising directions for future work highlighted by our results. The observed performance plateau, even on relatively similar domains, suggests improving the representation model both in terms of architecture and size. Since our model operates at the byte-level, hierarchical/multi-timescale extensions could improve the quality of representations for longer documents. The sensitivity of learned representations to their training domain could be addressed by training on a wider mix of datasets with better coverage of target tasks. Finally, our work encourages further research into language modelling as it demonstrates that the standard language modelling objective with no modifications is sufficient to learn high-quality representations.

\bibliography{example_paper}
\bibliographystyle{icml2017}

\end{document}